\documentclass[fleqn,10pt]{wlscirep}

\usepackage[utf8]{inputenc}
\usepackage[T1]{fontenc}
\usepackage[all]{nowidow}
\usepackage[english]{babel}
\title{Natural Language Processing in the Legal Domain}

\author[1,2,${\dagger}$,*]{Dirk Hartung}
\author[1,2,3,4,${\dagger}$]{Daniel Martin Katz}
\author[2]{Michael J. Bommarito}
\author[4]{Lauritz Gerlach}
\author[5]{Abhik Jana}
\author[1]{Jerrold Soh}
\affil[$^{\dagger}$]{These authors have contributed equally to this work and share first authorship.}

\affil[1]{Singapore Management University}
\affil[2]{CodeX, Stanford University, USA}
\affil[3]{Illinois Tech - Chicago Kent College of Law, USA}
\affil[4]{Bucerius Law School, Germany}
\affil[5]{IIT-Bhubaneswar}

\affil[*]{e-mail: dirkhartung@smu.edu.sg}

\begin{abstract}
We summarize the current state of the field of NLP \& Law with a specific focus on recent technical and substantive developments. To support our analysis, we construct and analyze a nearly complete corpus of nearly one thousand NLP \& Law related papers published between 2013--2024. Our analysis highlights several major trends. Namely, we document an increasing number of papers written, tasks undertaken, and languages covered over the course of the past decade. We observe an increase in the sophistication of the methods which researchers deployed in this applied context. Legal NLP is beginning to match not only the methodological sophistication of general NLP but also the professional standards of data availability and code reproducibility observed within the broader scientific community. We believe all of these trends bode well for the future of the field and point to an exciting next phase for the Legal NLP community. 
\end{abstract}

\begin{document}

\flushbottom
\maketitle

\thispagestyle{empty}

\section{Introduction}
Language is the coin of the legal realm. The careful drafting of documents and the analysis and interpretation of language are core activities undertaken by all participants in the world's legal systems. They produce large volumes of documents, and these documents are often difficult to understand and a source of increasing complexity.\cite{ruhl2017harnessing,katz2020complex,coupette2021measuring,vivo2024complexity,vivo2025complex} This has material consequences for access to justice.\cite{staudt2008all,ruhl2015measuring} Despite wide agreement on the resulting need for improvements in the delivery of justice,\cite{rhode2004access,wjp2019justicegap,prescott2017improving,susskind2019online,sandefur2020assessing}, scalable solutions to improve the quantity, quality, and accessibility of legal services remain elusive. Obstacles include cultural factors within the legal ecosystem \cite{susskind2017tomorrow,barton2017rebooting} and professional regulation.\cite{kobayashi2011law,hadfield2014cost,barton2018access} Yet, the primary \textit{technical} challenge that limits transformative technological solutions in law is the complexity of the legal language itself, which researchers have termed the ``natural language barrier'' \cite{mccarty_deep_2007,nazarenko_pragmatic_2021,ranta_end_2023}.
Training machines to understand legal language has been prohibitively challenging. Nonetheless, interest remained in exploring the potential of machines as force multipliers in processing legal texts. Scholars and commercial enterprises alike have explored the applicability of Natural Language Processing (NLP) technologies for legal use (Legal NLP). In academia, empirical legal studies increasingly rely on computer science methods to support analysis \cite{livermore2019law,fagan2020natural,kolt2022predicting}. In industry, domain-specific NLP tools designed specifically for lawyers currently sell at a premium to offerings for the general population. As a result, there have been attempts to embed Legal NLP modules into several legal practice applications \cite{dale2019law,engstrom2020legal,bommarito2021lexnlp}, ranging from research tools and litigation outcome prediction to drafting support and compliance risk assessment. Overall, despite several noteworthy attempts, the performance of both academic and commercial Legal NLP applications has conventionally been confounded by computer`s inability to accurately and reliable process legal language.
\newpage
\noindent
Meanwhile, the  technical literature beyond law has witnessed major gains in the past 15 years, building on foundational advances in neural network research,\cite{Rumelhart1986}\textsuperscript{,}\cite{LeCun2015}, and leading up to significant performance breakthroughs with large language models (LLMs). While early neural NLP papers were built on word embeddings,\cite{mikolov2013distributed,pennington2014glove,peters2018deep} the latest wave of transformer-based LLMs \cite{vaswani2017attention} allows for clever manipulation of the attention mechanism so that training tasks can be scaled through more effective parallelization.\cite{tay2022efficient} Despite some critique,\cite{bender2021dangers} successive iterations of increasingly large transformer-based language models have delivered remarkable results. \cite{kenton2019bert,brown2020language,zaheer2020big,scao2022bloom,thoppilan2022lamda}. LLMs are increasingly characterized as general purpose models that could be, and are being, used by lawyers. Nonetheless, the transferability of general models to specific complex domains such as law remains an open question: general models have shown real progress on legal tasks\cite{zheng2021does,chalkidis2022lexglue,bommarito2023gpt,nay2023large,katz2024gpt}, but there remain reasons\cite{Huang2022,wu2022promptchainer,zhou2022large} to believe that some combination of domain-specific pre-training, prompt and context engineering, and other model tuning efforts yield better results across many substantive use cases. In any event, rapid technological progress coupled with the expansion of digitized legal data means the prospect of deploying cutting-edge NLP techniques in law has risen sharply in recent years while costs fall. NLP models have thus proliferated in legal research, practice, and education, as evidenced by a growing set of survey and review publications focused on methods, use cases, challenges, and tasks.\cite{ariai2025natural,dehghani2025large,hu2026evaluation} 
To inform these efforts and provide a roadmap for various interdisciplinary scholars, this review paper documents and summarizes emergent trends in Legal NLP. Although there have been efforts to characterize some developments in the field, including some important analyses of Legal NLP alone \cite{fagan2020natural,zhong2020does} and discussions situated in the broader context of legal informatics,\cite{ashley2017artificial,katz2021legal} we believe a more holistic treatment of the current state of Legal NLP is justified. Our systematic survey covers Legal NLP papers published from 2013 to 2024. Notably, our initial review covering publications from 2013 to 2022 had been circulated as a pre-print ([Citation removed to avoid de-anonymization). Significant developments in (legal) LLMs from 2022 onwards, however, motivated us to collect additional data in order to provide an updated picture presented below.

% To inform these efforts and provide a roadmap for various interdisciplinary scholars, this review paper documents and summarizes emergent trends in Legal NLP. Although there have been efforts to characterize some developments in the field, including some important analyses of Legal NLP alone \cite{fagan2020natural,zhong2020does} and discussions situated in the broader context of legal informatics,\cite{ashley2017artificial,katz2021legal} we believe a more holistic treatment of the current state of Legal NLP is justified. Our systematic survey covers Legal NLP papers published from 2013 to 2024. Notably, our initial review covering publications from 2013 to 2022 had been circulated as a pre-print \cite{katz2023natural}. Significant developments in (legal) LLMs from 2022 onwards, however, motivated us to collect additional data in order to provide an updated picture presented below.

\section{Building the Corpus of Legal NLP Papers}
\label{sec:2}
\begin{figure}[htbp]
  \centering
  \includegraphics[width=0.75\linewidth]{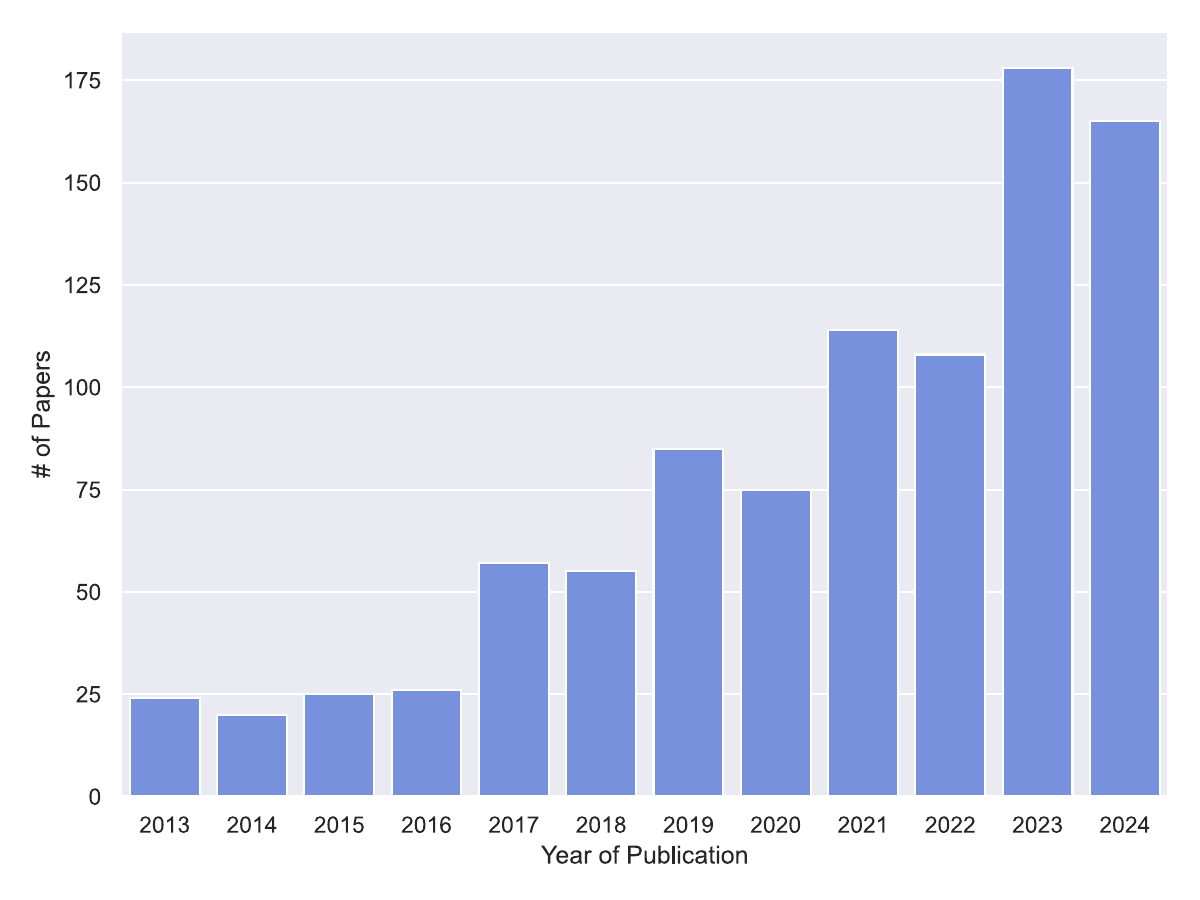}
  \caption{Number of Legal NLP Papers over Time}
  \label{fig:1}
\end{figure}

To identify the corpus of Legal NLP papers, we began by reviewing the proceedings of both general NLP conferences (e.g.\ ACL, NAACL, EMNLP, EACL, etc.) and specialty Legal NLP gatherings (ICAIL, Jurix, MLLD, NLLP etc). A total of 21 publication outlets were included. Next, we performed iterative queries on major search engines and publication databases (such as Google Scholar, SSRN and arXiv). Finally, we performed `snowball sampling' by manually traversing the citation graph of key publications to look for additional Legal NLP papers not already identified. Absent extraordinary circumstances, we restricted our corpus to include only peer-reviewed scientific journals, technical conference proceedings, and otherwise technically-oriented pre-prints. We excluded essays, commentaries, blog posts, social critiques, or otherwise non-technical or empirical publications. 
With the help of several legally trained research assistants, we reviewed the initial long list of papers manually to determine if a given paper indeed fell within \textit{Legal} NLP. Qualitative judgment is required because law is an intellectual domain which sits at the intersection of the humanities and social sciences; its boundaries are famously open-textured. To illustrate, many papers apply NLP to analyze documents related to financial instruments. Although the difference between these financial documents and legal documents can often be marginal, the focus and audience of these papers is typically very different. Therefore, we excluded these due to their limited connection to the legal domain. Our guiding principle was to include any technical paper whose target audience was a legal scholar or practitioner. We further focused the `NLP' aspect on papers adopting statistical, machine learning, corpus linguistics and similar approaches (these terms too are not always well-defined). Thus, works solely focused on symbolic logic and formal methods\cite{sergot_british_1986,bench-capon_logic_1987,palmirani_legal_2018} were excluded. This was unless the work incorporated a significant statistical machine learning component to form an effectively neurosymbolic approach, for instance, by using NLP models to convert legal texts to logical form\cite{ranta_end_2023} or generate case-based arguments\cite{gray_2025}.
The final dataset includes 932 papers and can be accessed at [insert reference to final data repository after acceptance, for the review process please see https://drive.google.com/drive/folders/1ljWATLKNh7NeH47CaVHzAdY7kJGvcU8F?usp=sharing which contains both the raw paper pdfs and the results of the data extraction and analysis. Please use Chrome in incognito mode to avoid de-anonymization]. Figure \ref{fig:1} charts the volume of papers published over those 12 years. We observe a significant growth in total publications. Interestingly, a periodic variation pattern emerges as every other year sees a significant increase followed by a slight decrease, which we attribute to certain conferences (e.g. ICAIL) happening only in even-numbered years during our period of observation. We find nearly a 7-time increase between 2013 and 2024. The growth is drastically accelerating as the total number of papers per year doubles roughly every three years. More than a third of our corpus was written after 2022, which coincides with the wide availability of LLMs. 

The remaining sections of this paper report trends that we have observed in this dataset: growth across tasks (section \ref{sec:3}); evolution of methods (section \ref{sec:4}); diversity of languages (section \ref{sec:5}); reproducibility and data availability (section \ref{sec:6}); and citation counts (section \ref{sec:7}).

\section{The Growth of Legal NLP Across Various Tasks}
\label{sec:3}
Law regulates all sectors of society. The topic space covered is vast, extending across legal subjects such as contract, tort, or constitutional law and practice areas from admiralty to trust law. Within each area, lawyers undertake a diversity of tasks. Popular legal task schemata (\textit{e.g.} , UTBMS, SALI) typically further decompose legal workstreams into constituent subtasks.\cite{bartolo2019pre,constantinou2020detecting} For example, within a given legal matter, a legal professional might review documents, draft documents, conduct research, give arguments in court, negotiate with counterparties, and explain the law to their clients. This task diversity is reflected across our collection of nearly 1000 papers in which scholars seek to apply Legal NLP techniques to augment, assist, or even replace lawyers and paralegals.

Parallel to the above lawyer-centric perspective, Legal NLP tasks can also be examined from an engineering angle. Relevant engineering tasks include machine summarization, generation, classification, retrieval, etc. These perspectives often intersect. Consider a paper that explores the automated extraction of keywords from a patent claim in order to support the search for prior art (i.e. other relevant patents).\cite{rossi2018query} This paper can be characterized as a patent paper from a substantive law perspective and as an information retrieval type paper from an engineering perspective. Recognizing that this dualism affects the vast majority of papers in our corpus, we will – for the purposes of our analysis –privilege the engineering perspective on Legal NLP tasks.

\begin{table*}[hbt!]
\centering
\begin{tabular}{|l|l|l|}
\hline
 & \multicolumn{1}{c|}{\textbf{Task}} & \textbf{Examples / Description}                                                       \\ \hline
1 & \textbf{Machine Summarization}   & Abstractive/Extractive Summaries of Legal Documents                                               \\ \hline2 & \textbf{Pre-Processing}      & Annotation, Anonymization, Translation                                               \\ \hline
3 & \textbf{Classification}      & \begin{tabular}[c]{@{}l@{}}Outcome Prediction, Legal Area Classification, Topic Modeling
\end{tabular} \\ \hline
4 & \textbf{Information Retrieval}   & Legal Question Answering, Document Similarity, Document Retrieval                                           \\ \hline
5 & \textbf{Information Extraction}  & Labeling, Text Extraction, Event Extraction                                                   \\ \hline
6 & \textbf{Text Generation}      & Automated Drafting of Legal Documents                                                \\ \hline
7 & \textbf{Resources}         & Benchmarks, Evaluations, Taxonomies, Ontologies, Datasets, Code Libraries                                          \\ \hline
\end{tabular}
\caption{\label{tab}A Taxonomy of Engineering Tasks in Legal NLP}
\end{table*}

\begin{figure}
  \centering
  \includegraphics[width=0.75\linewidth]{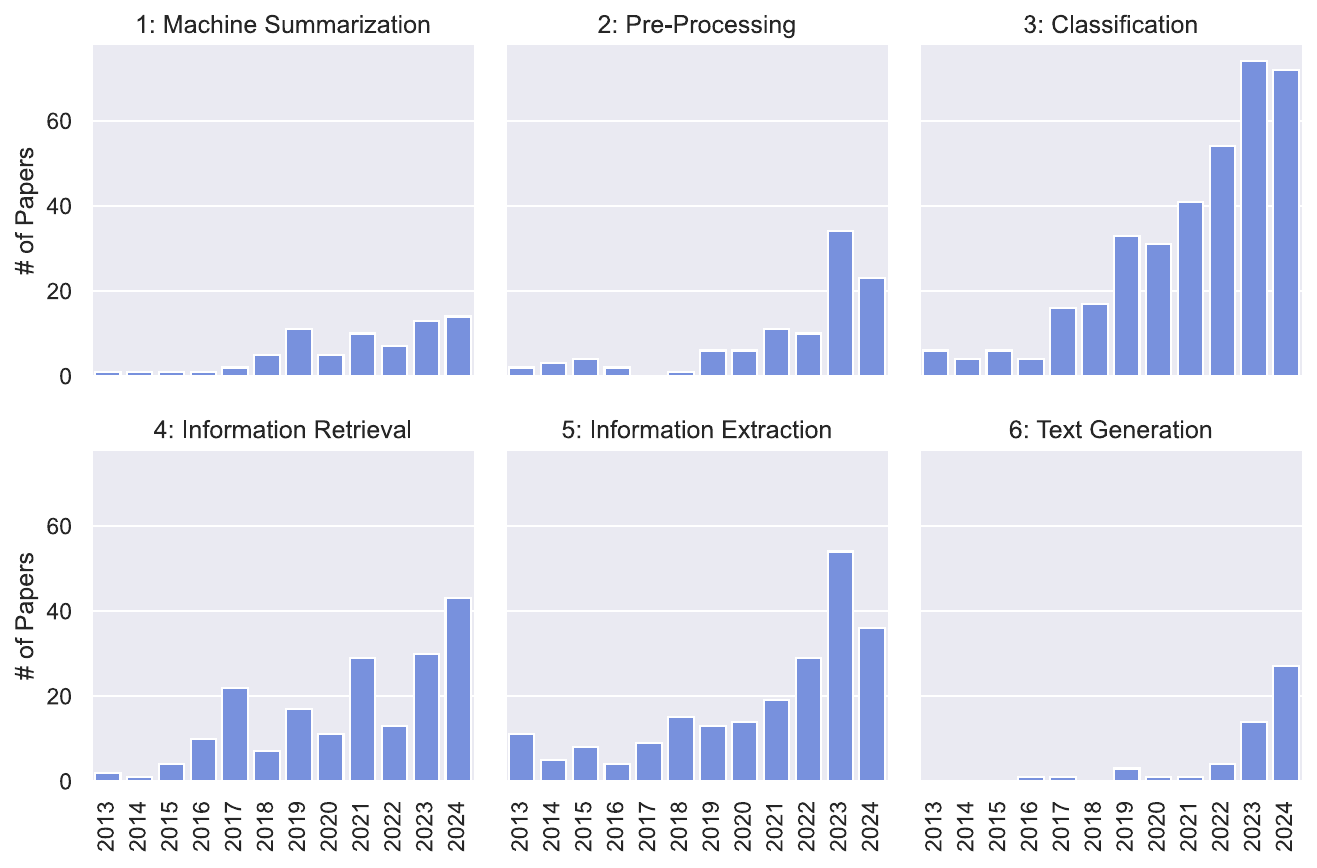}
  \caption{Legal NLP Tasks over Time}
  \label{fig:2}
\end{figure}

\noindent 

While certainly not the only way one could subdivide the space, Table \ref{tab} offers such an engineering-centric taxonomy of Legal NLP tasks based on an initial subject-matter expert labeling of all papers in our corpus. Using this taxonomy, we qualitatively reviewed each of the papers and determined the relevant engineering categor(ies) for the work contained therein. Although many papers fit squarely within one particular category, an increasing number of papers contain two or more engineering tasks. Figure \ref{fig:2} reflects the temporal distribution of the papers by engineering task for tasks 1--6. 
\begin{figure}
  \centering
  \includegraphics[width=1\linewidth]{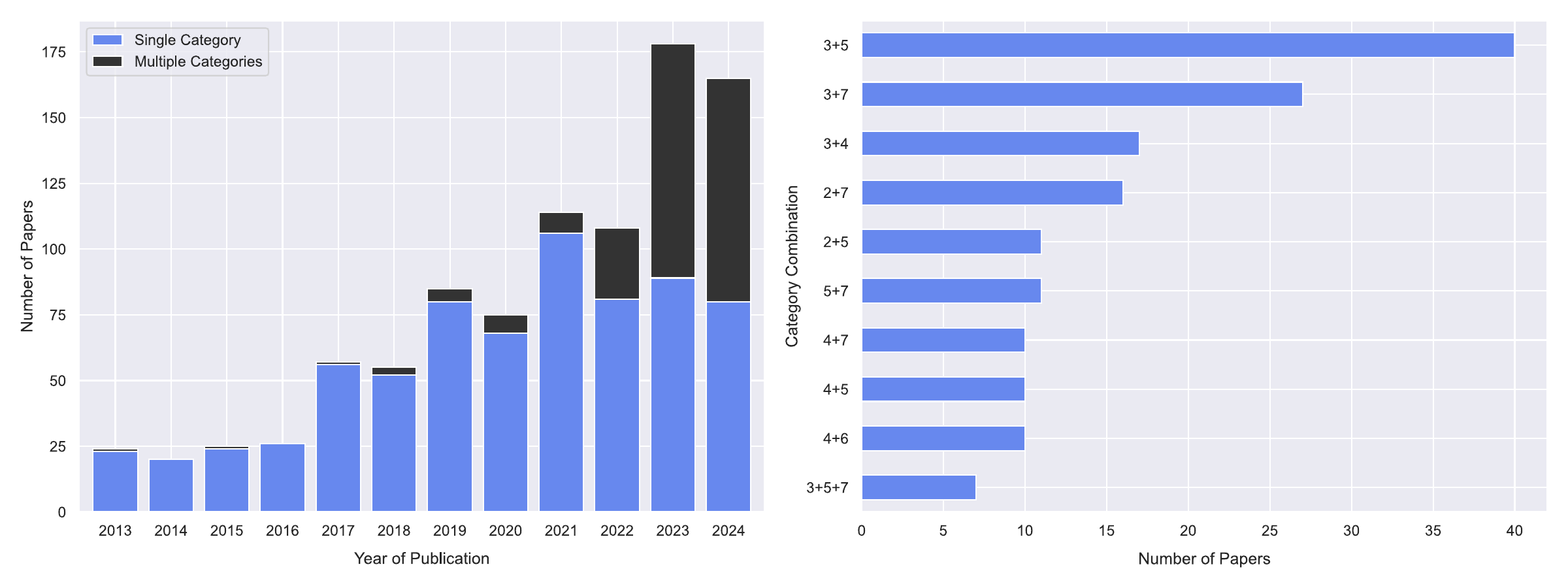}
  \caption{Increasing sophistication of paper categories}
  \label{fig:3}
\end{figure}

\begin{figure}[htbp]
  \centering
  \includegraphics[width=0.75\linewidth]{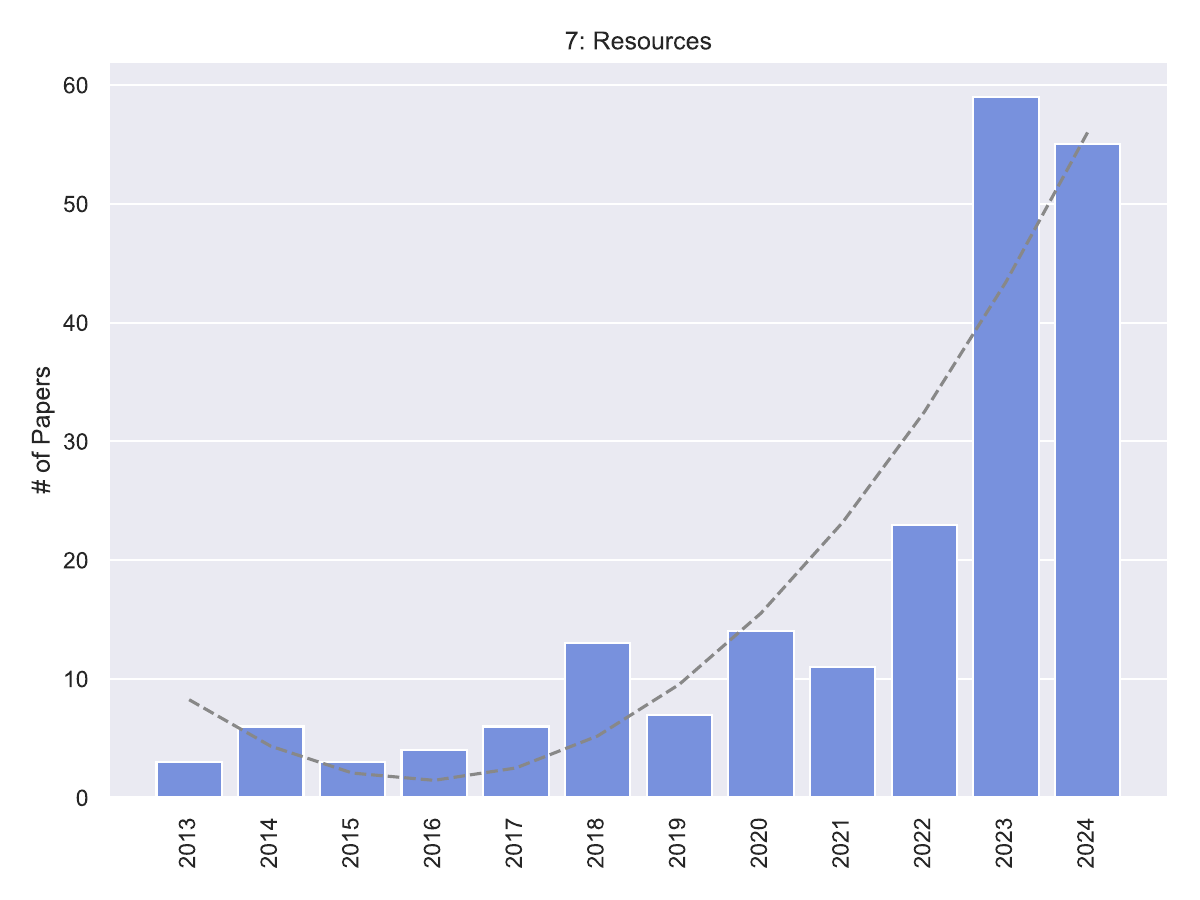}
  \caption{Resources papers over time}
  \label{fig:4}
\end{figure}

\noindent
As can be seen in Figure \ref{fig:3}, the emergence of multi-category papers is a rather recent phenomenon. Before 2018, virtually no Legal NLP paper was categorised within more than one task category. Overlaps began to appear around 2018 and 2019, which notably coincides with the advances in LLM such as GPT and BERT \cite{kenton2019bert,brown2020language}. We also saw a sharp rise in 2023 and 2024, where more than half of the papers documented fell with multiple categories. The right-side panel of Figure \ref{fig:3} visualises how frequently task categories (denoted by their IDs from Table \ref{tab}). As these combinations occur in repeating patterns and characterize a new, technically more complicated type of research likely enabled by the ubiquity of LLMs, we call them motifs. Most motifs combine a classification task with one other task such as information extraction and retrieval as well as pre-processing and evaluation. However, some papers combine even more categories.
Subdividing the papers by categories yields another interesting result: the general trend from an overall increase in papers is carried through all categories, albeit unequally distributed. Summarization and Classification show a change of pace around 2018. Extraction and Information Retrieval, on the other hand, show fairly stable interest over time. Both Resources and Pre-Processing papers increase over time with the general stable increase of papers until 2022 and sudden increase thereafter.
Some applications, such as Machine Summarization and Text Generation, had previously seen little activity in the earlier parts of the period under observation. In an earlier version of this paper, we therefore expected a marked increase in the volume of these categories.\cite{katz2023natural} This has been true for Text Generation, which has grown remarkably since 2022. Summarization, in turn, has grown, too, but not to that extent. However, in general, the advent of LLMs is likely to be reflected in this trend.
Finally, the last two years have witnessed the ascension of the Resources category, which we display separately in Figure \ref{fig:4}. This category primarily contained taxonomies, onotologies, benchmarks, datasets, and code libraries in the past. The recent increase is driven by a large number of evaluation papers. These publications typically employ different LLMs to handle legal tasks from a specific substantive and/or practice areas and evaluate the results against human subject-matter experts or existing benchmarks in a variety of languages and domestic laws. The rise of such papers points to the proliferation of NLP tools in legal research, since they represent a shift from purely technical to increasingly legal evaluations.

\section{The Evolution of Methods in Legal NLP}
\label{sec:4}
Early work in NLP can be traced to various rules-based systems which were either proposed or implemented.\cite{chomsky1957syntactic}\textsuperscript{,} \cite{schank1973margie}\textsuperscript{,}\cite{lehnert1977conceptual} A long-term significant increase in computing power,\cite{shalf2020future} and the vast decline in the cost of data storage,\cite{gupta2014economic} taken together with algorithmic advances,\cite{krichen2024performance,vaswani2017attention} saw the field of NLP transformed from its rules-based AI origins into a data-driven field. This statistical turn, which began in the 1990's, had given way to the `neural era' within the general field of NLP in the 2010s.\cite{zhou2020progress} Recently, an extension of these developments and a rapidly growing body of available training data has lead to even more capable models known as Large Language Models (LLM).\cite{minaee2024large}
Rooted in Legal Informatics, the field of Artificial Intelligence and Law has a long tradition starting several decades ago, when rules-based and logic-based artificial intelligence was dominant. During our period of observation, data- and machine-learning-based approaches have become more dominant, following similar developments in computer science and engineering.\cite{governatori2022thirty}  To study this shift, we selected a set of key words to track the methods applied within each paper. We follow an established development path from symbolic and statistical NLP via distributed representations, deep architecture, pre-training, and finally scale.\cite{goldberg2016primer,qiu2020pre,zhao2023survey} We have chosen a set of 25 terms representative of techniques within this evolution from tf-idf and svm via word2vec, rnn, and lstm to transformer, llm, and rag.

\begin{figure*}[htbp]
  \centering
  \includegraphics[width=0.76\linewidth]{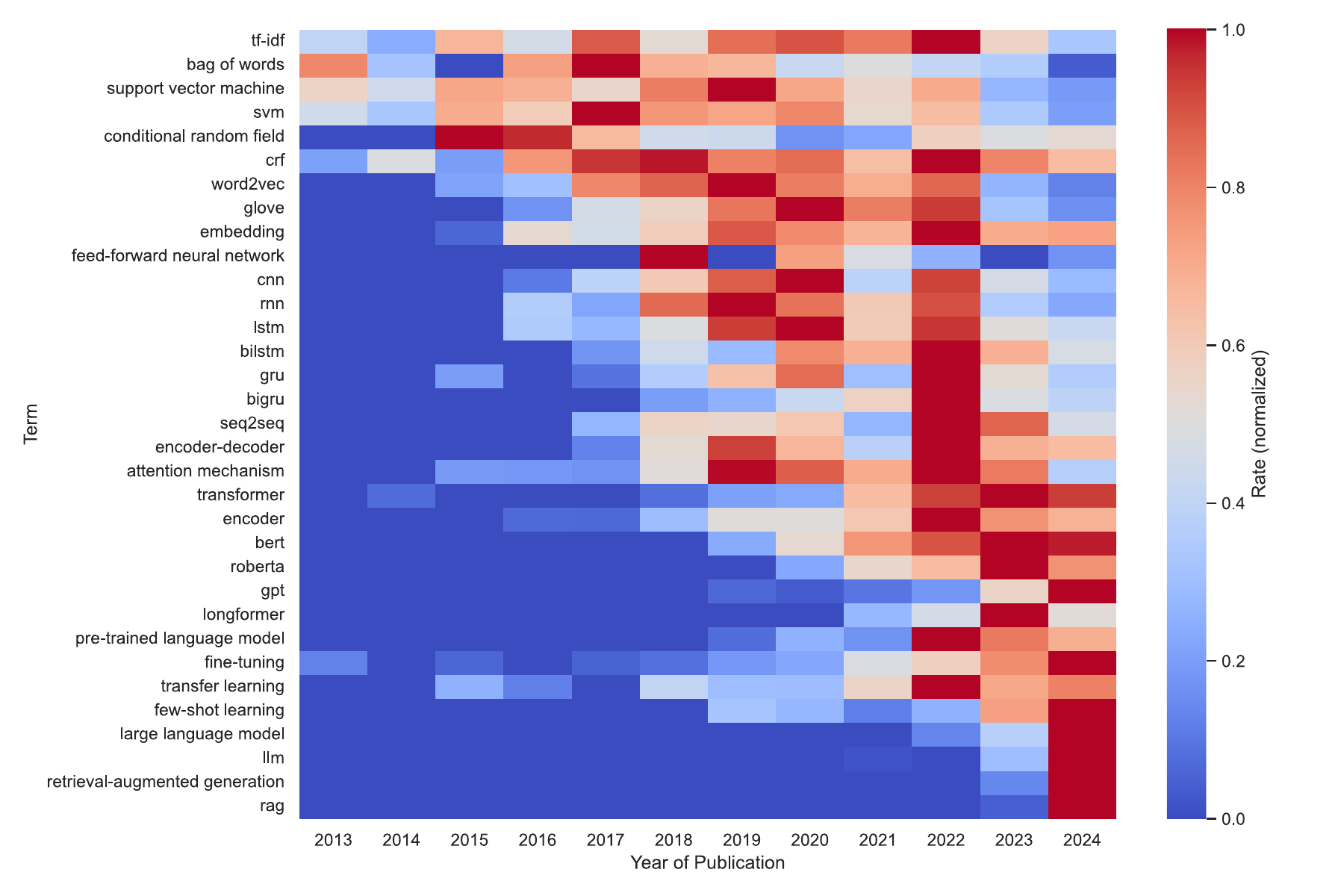}\\ 
  \caption{Relative Rate of Term Usage over Time. Normalization is per-term relative to the maximum annual rate of mentioning papers.}
  \label{fig:5} 
\end{figure*}

\noindent
Figure \ref{fig:5} offers a normalized time series of key phrases on a yearly basis. Some fundamental methods such as tf-idf and other bag of words approaches appear distributed across the entire time, possibly as they are used for baseline comparisons. Other terms, such as terms related to LLMs, only appear towards the end of observation period. Overall, individual eras of methods can be traced rather well in our data set. Most methods seem to reach peak popularity in the field of Legal NLP some time after they were first published; however, that delay tends to get shorter for newer methods. In total, the data show persistent progress within the field and general trends towards cutting-edge models at any given point in time.

\section{The Diversity of Languages}
\label{sec:5}
Due to the Law's territoriality, picking the language of the corpus is a fundamental decision at the core of each paper. English is historically well-positioned as the standard language of both computer science scholarship and international (business) law, and thus unlocks vast datasets from countries as diverse as the United States, the United Kingdom, India, Singapore, and Australia. For this reason, we expected English to dominate the distribution of languages in our papers, with other large languages such as Chinese, German, and French to be the nearest contenders. To interrogate this further, we manually assigned language label(s) to all papers within our dataset. This was based primarily on the language of the legal texts analyzed within each paper. Some papers received more than one language label, for example when they dealt with German and American statutes.\cite{katz2020complex} Furthermore, we had to make a choice for papers with multiple languages (about 9\%), e.g. from multilingual jurisdictions and the European Union whose legal documents often appear in a variety of languages. In these instances, we assign one label for each language version used for the actual analysis and a dedicated \textit{EU} label. As a result, EU law is rather prominent in the third place with just under 8\% of all articles.
Our initial hypothesis was mostly correct: the most popular language in the entire corpus is English (54\%). The next most frequent is Chinese (10\%), which displays particularly interesting temporal dynamics as described below. Other languages such as German, French, Portuguese, Japanese and Italian each make up between 3\% and 5\%, likely reflecting their digitally available legal corpora and disproportionately large research communities in both computer science and law. 

\begin{figure}[htbp]
  \centering
  \includegraphics[width=0.75\linewidth]{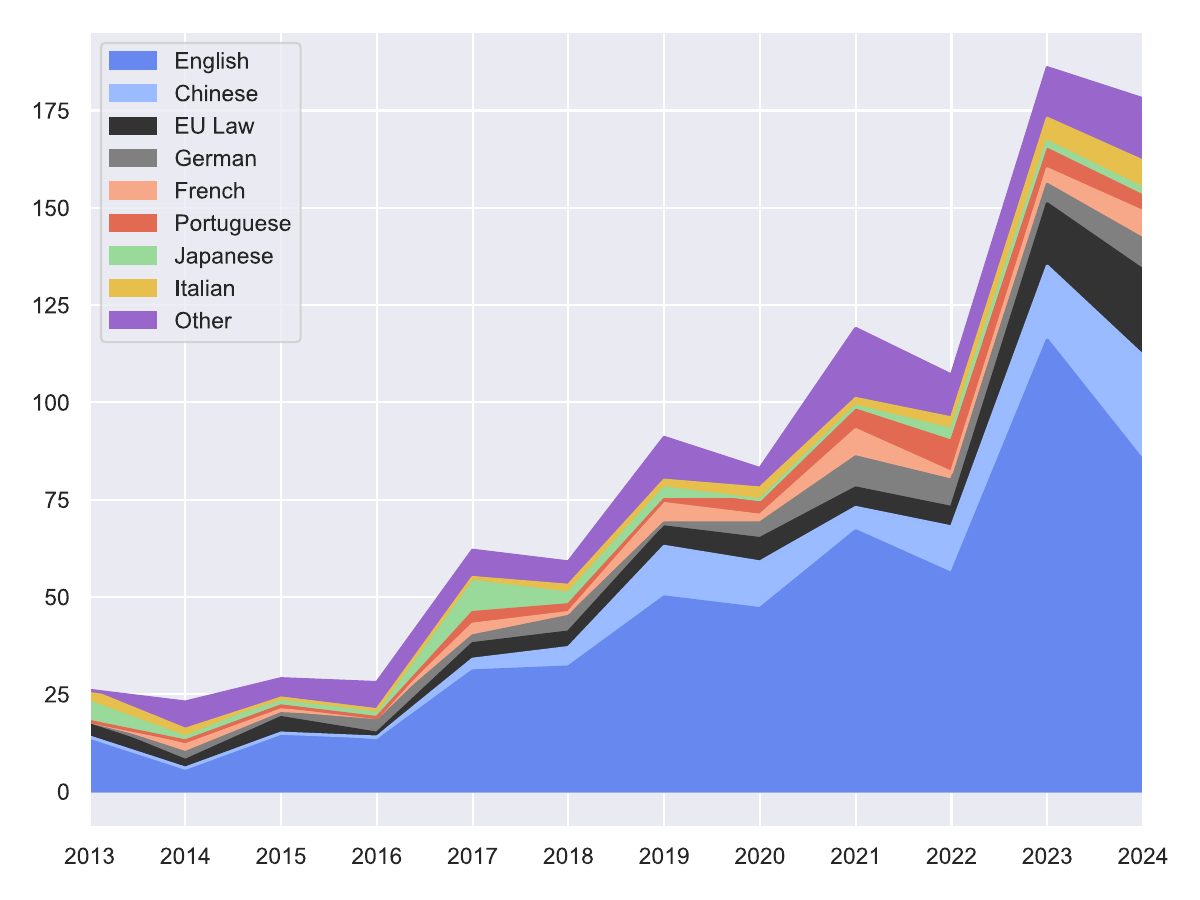}
  \caption{Number of publications by language of the analyzed legal corpus over time, with EU law shown separately as it is published in a variety of languages.}
  \label{fig:6}
\end{figure}

\noindent
Figure \ref{fig:6} shows a stack plot of the time series of papers by language. Observing the distribution of the five most common languages over time per year, the proportion of English-language papers remains roughly constant, while papers analyzing Chinese-language corpora increase substantially and more strongly in the later years, particularly after 2023. Additionally, Figure \ref{fig:6} reveals that the overall diversity of languages has increased, with the sum of less common languages increasing pertinently over time (cf. the `other' band in Figure \ref{fig:6}). As a word of limitation, it is important to note that we only surveyed papers that were themselves written in English, which introduces a relevant bias.\cite{amano2023manifold} Therefore, given that significant scholarship on non-English languages is likely published only in those languages, the results offered in Figure \ref{fig:6} should be considered as lower bound estimates for linguistic diversity.

\section{Reproducibility and Data Availability}
\label{sec:6}
Fueled by an extensive crisis in some disciplines, reproduction and reproducibility have become an increasing concern not only for Legal NLP but also for the broader scientific community.\cite{baker2017reproducibility,munafo2017manifesto,ivie2018reproducibility} Beyond the important task of verifying existing results, transparent and replicable output can help accelerate the pace of additive innovation. If not just the data, but also the code of existing work is made available, this significantly facilitates and accelerates subsequent research endeavors.
For each of the nearly one thousand papers, we comprehensively evaluated the nature of replication resources made available by the respective authors. In judging the reproducibility of a given paper, many different aspects can be considered, including the availability of datasets, corpora, models, and (easy-to-follow) documentation. We categorize all the papers in our corpus into one of three classes. \textit{Class I} contains those papers for which either no proper links to any resources are mentioned or the given links do not or no longer work. For this class, it would be extremely burdensome and bordering impossible to reproduce the results. \textit{Class II} papers provide partial resources to support replication / implementation. The nature of the incomplete element varies. Some papers provide only data sources, whereas for some papers, only code is available, often with little or no documentation. Therefore, the frameworks or the results presented in this class of papers cannot be reproduced straightforwardly without the material assistance of the authors. Finally, \textit{Class III} papers offer a well-organized repository of resources (both code and data) with proper documentation.

\begin{figure}[htbp]
  \centering
  \includegraphics[width=1\linewidth]{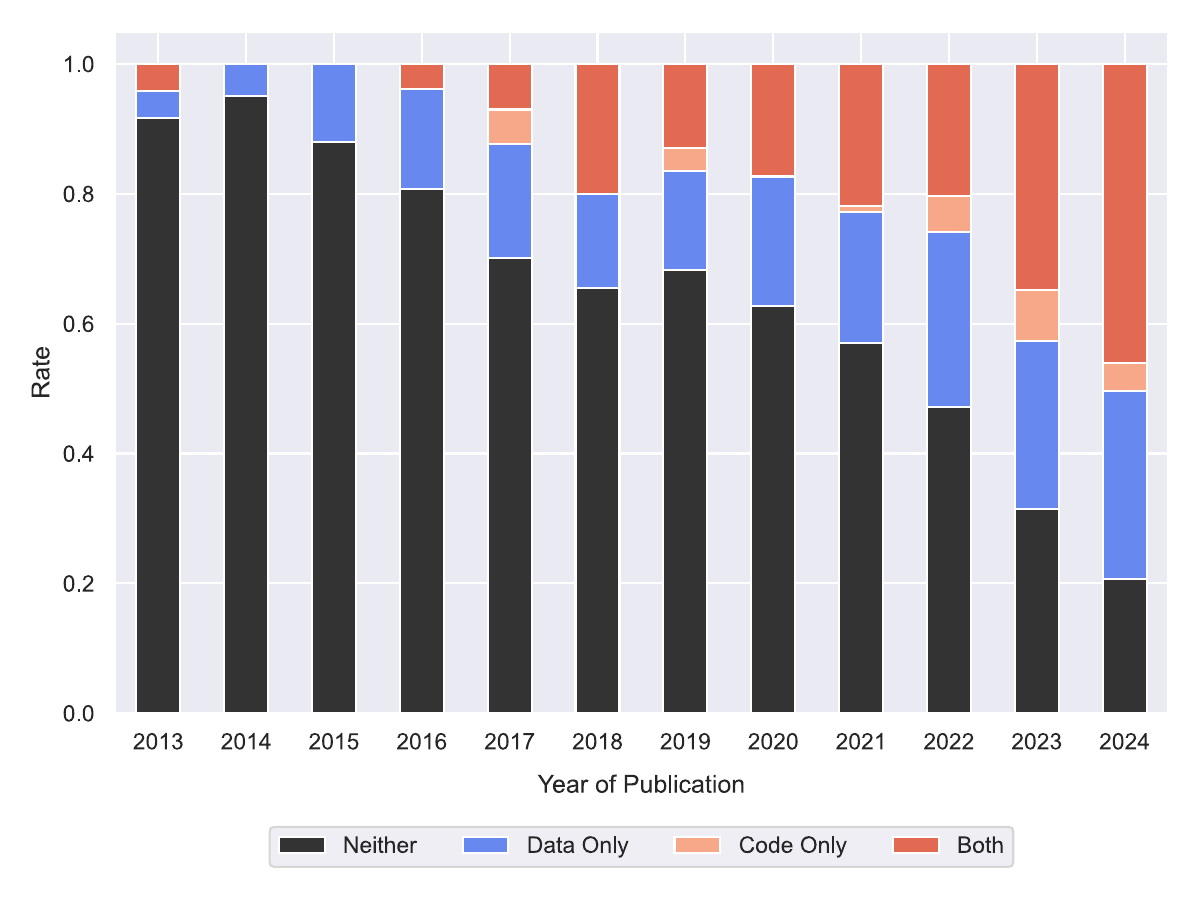}
  \caption{Replication Material Availability as a Function of Time}
  \label{fig:fig7}
\end{figure}

\noindent
Based on this classification, we analyze the availability of resources from two different perspectives. In Figure \ref{fig:fig7}, we investigate whether the authors of Legal NLP have over time become more inclined to make resources publicly available. The results for the period under review are promising. The proportion of \textit{Class I} papers has decreased significantly from 92\% (2013) to just 20\% (2022) whereas \textit{Class III} papers increased noticeably from 4\% (2013) to 46.1\% (2022). These statistics highlight an increased commitment to reproducibility within the Legal NLP community. Examined over a longer time period, the period up to 2016 had rather low reproducibility while 2017--2022 saw steady growth that strongly accelerated in 2023 and 2024. At this point in time, more than half the papers make their code available and 3/4 papers provide a data set. We interpret these results as an indication of increasing rigor and professionalism within the Legal NLP community and, at the same time, see them as an effect of a move towards open availability of legal data in many jurisdictions.
Read in conjunction with our impact analysis below in Section \ref{sec:7}, it becomes clear that the community rewards adherence to best practices: papers containing datasets (${\sim}65\%$), code (${\sim}30\%$) or both (${\sim}63\%$) gain considerably more citations on average than those without. However, an analysis of variance shows that while the availability of either dataset or code significantly boosts citations, provided that both does not have an additional effect. In addition, while the statistical effect is significant, the effect size is rather negligible, suggesting that many other factors (including obvious ones such as age of the publication) drive citations more strongly than resource availability alone.

\section{Legal NLP Impact Analysis}
\label{sec:7}
As highlighted in Section \ref{sec:2}, the field of Legal NLP is growing strongly. To study the dynamics of such growth in more detail, we collected citation data. This can provide a useful perspective on key topics, papers, and researchers as well as impact within the field. While citations are a noisy measure of `quality' and necessarily a snapshot, they are a crowd-sourced measure of prominence within the scientific community.\cite{ioannidis2016multiple,tahamtan2019citation} 
Using the Google Scholar Cite API via SerpApi, we automated the collection of citation data for all papers and subsequently reviewed the results for quality control purposes. The presented citation counts reflect the current status as of September 2025. Table \ref{tab:most_cited} offers a list of the most cited papers by total citations while the full ranking is available online. Even a cursory review reflects the wide variety of substantive topics and jurisdictions. Collectively, these papers represent a good initial starting point for those interested in learning more about the emerging field of Legal NLP. 
Overall, while citation counts vary widely, an expected bibliometric skew\cite{brzezinski2015power} towards the top papers is present within the field of Legal NLP, too. Publication age was a weak but significant contributing factor to citation count ($\rho$ = 0.281), suggesting a more sustained citation accumulation than bibliometric burst-and-decay models would predict.\cite{wang2009effect,tahamtan2016factors}. This could be explained at least in part by the fact that the longer a paper exists, the more opportunities it has to be cited and the fact that – unlike in some faster moving fields with more publications than during the early days of our field – Legal NLP results remain relevant for longer periods of time. To account for this, we additionally provide the age-normalized rank in parentheses in Table \ref{tab:most_cited}.

\begin{table*}
\centering
  \resizebox{\textwidth}{!}{
  \begin{tabular}{l|p{10cm}|p{7cm}|l}
  \toprule
  \bf Rank & \bf Title & \bf Authors & \bf Year\\
  \midrule
\textbf{1} (4)    & LEGAL-BERT: The Muppets straight out of Law School &    Chalkidis, Fergadiotis, Malakasiotis, Aletras, Androutsopoulos &    2020 \\
\midrule
\textbf{2} (7)    & Predicting judicial decisions of the European Court of Human Rights: a Natural Language Processing perspective &    Aletras, Tsarapatsanis, Preoțiuc-Pietro, Lampos &    2016 \\
\midrule
\textbf{3} (1)    & GPT-4 passes the bar exam &    Katz, Bommarito, Gao, Arredondo &    2024 \\
\midrule
\textbf{4} (24)    & Using machine learning to predict decisions of the European Court of Human Rights &    Medvedeva, Vols, Wieling &    2020 \\
\midrule
\textbf{5} (5)    & Neural Legal Judgment Prediction in English &    Chalkidis, Androutsopoulos, Aletras &    2019 \\
\midrule
\textbf{6} (28)    & How Does NLP Benefit Legal System: A Summary of Legal Artificial Intelligence &    Zhong, Xiao, Tu, Zhang, Liu, Sun &    2020 \\
\midrule
\textbf{7} (6)    & Legal Judgment Prediction via Topological Learning &    Zhong, Guo, Tu, Xiao, Liu, Sun &    2018 \\
\midrule
\textbf{8} (3)    & LexGLUE: A Benchmark Dataset for Legal Language Understanding in English &    Chalkidis, Jana, Hartung, Bommarito, Androutsopoulos, Katz, Aletras &    2022 \\
\midrule
\textbf{9} (2)    & LEGALBENCH: A Collaboratively Built Benchmark for Measuring Legal Reasoning in Large Language Models &    Guha, Nyarko, Ho, Ré, Chilton, Narayana, Chohlas-Wood, Peters, Waldon, Rockmore, Zambrano, Talisman, et al. &    2023 \\
\midrule
\textbf{10} (8)    & Learning to Predict Charges for Criminal Cases with Legal Basis &    Luo, Feng, Xu, Zhang, Zhao &    2017 \\
\midrule
\textbf{11} (9)    & Lawformer: A pre-trained language model for Chinese legal long documents &    Xiao, Hu, Liu, Tu, Sun &    2021 \\
\midrule
\textbf{12} (15)    & Few-Shot Charge Prediction with Discriminative Legal Attributes &    Hu, Li, Tu, Liu, Sun &    2018 \\
\midrule
\textbf{13} (10)    & When Does Pretraining Help? Assessing Self-Supervised Learning for Law and the CaseHOLD Dataset of 53,000+ Legal Holdings &    Zheng, Guha, Anderson, Henderson, Ho &    2021 \\
\midrule
\textbf{14} (11)    & CUAD: An Expert-Annotated NLP Dataset for Legal Contract Review &    Hendrycks, Burns, Chen, Ball &    2021 \\
\midrule
\textbf{15} (13)    & Natural Language Processing to Identify the Creation and Impact of New Technologies in Patent Text: Code, Data, and New Measures &    Arts, Hou, Gomez &    2021 \\
\midrule
\textbf{16} (17)    & CLAUDETTE: an automated detector of potentially unfair clauses in online terms of service &    Lippi, Pałka, Contissa, Lagioia, Micklitz, Sartor, Torroni &    2019 \\
\midrule
\textbf{17} (16)    & Text summarization from legal documents: a survey &    Kanapala, Pal, Pamula &    2017 \\
\midrule
\textbf{18} (12)    & A comparative study of automated legal text classification using random forests and deep learning &    Chen, Wu, Chen, Lu, Ding &    2022 \\
\midrule
\textbf{19} (25)    & Deep learning in law: early adaptation and legal word embeddings trained on large corpora &    Chalkidis, Kampas &    2018 \\
\midrule
\textbf{20} (21)    & Exploring the Use of Text Classification in the Legal Domain &    Șulea, Zampieri, Malmasi, Vela, Dinu, Genabith &    2017 \\
\midrule
\textbf{21} (69)    & TechNet: Technology Semantic Network Based on Patent Data &    Sarica, Luo, Wood &    2020 \\
\midrule
\textbf{22.5} (27)    & JEC-QA: A Legal-Domain Question Answering Dataset &    Zhong, Xiao, Tu, Zhang, Liu, Sun &    2019 \\
\midrule
\textbf{22.5} (75)    & Distinguish Confusing Law Articles for Legal Judgment Prediction &    Xu, Wang, Chen, Pan, Wang, Zhao &    2020 \\
\midrule
\textbf{24.5} (79)    & BERT-PLI: Modeling Paragraph-Level Interactions for Legal Case Retrieval &    Shao, Mao, Liu, Ma, Satoh, Zhang, Ma &    2020 \\
\midrule
\textbf{24.5} (18)    & LawBench: Benchmarking Legal Knowledge of Large Language Models &    Fei, Shen, Zhu, Zhou, Han, Huang, Zhang, Chen, Yin, Shen, Ge, Ng &    2024 \\
\midrule
\textbf{26} (23)    & ILDC for CJPE: Indian Legal Documents Corpus for Court Judgment Prediction and Explanation &    Malik, Sanjay, Nigam, Ghosh, Guha, Bhattacharya, Modi &    2021 \\
\midrule
\textbf{27} (33)    & Legal Judgment Prediction via Multi-Perspective Bi-Feedback Network &    Yang, Jia, Zhou, Luo &    2019 \\
\midrule
\textbf{28} (34)    & Extracting Contract Elements &    Chalkidis, Androutsopoulos, Michos &    2017 \\
\midrule
\textbf{29} (36)    & A Comparative Study of Summarization Algorithms Applied to Legal Case Judgments &    Bhattacharya, Hiware, Rajgaria, Pochhi, Ghosh, Ghosh &    2019 \\
\midrule
\textbf{30} (14)    & Copyright Violations and Large Language Models &    Karamolegkou, Li, Zhou, Søgaard &    2023 \\
\bottomrule
\end{tabular}
}
\caption{Most cited Papers in Legal NLP published since 2013 by total citation counts and in parentheses after normalization.}
\label{tab:most_cited}
\end{table*}

\section{An Interactive Living Survey}
\label{An Interactive Living Survey}
A survey spanning more than a decade of global research is an enormous task, even in an emerging field such as Legal NLP. The sheer amount of continuously created new information leads to a significant risk of overburdening the community, resulting in scientific findings being overlooked or lost. 
Even at the beginning of the period under review, the growth rate of publications in our field (see infra \ref{fig:1}) made it difficult to identify all relevant publications. As the number of new publications over time grew and the methodical and linguistic diversity increased, the task has become more and more challenging. Similar developments in science in general \cite{bornmann2015growth} have inspired a series of innovative attempts to solve the problem of scientific knowledge discovery \cite{kumar2025large} and education \cite{thoppilan2022lamda} by using large language models to support both research \cite{beltagy2019scibert} and writing of scientific publications \cite{taylor2022galactica} albeit with limited success and apparent upper bounds despite large numbers of parameters.\cite{hong2023diminishing} Given technical limitations and employing an established design mechanism from machine learning, \cite{wu2022survey,wang2022human} we suggest a hybrid, human-in-the-loop approach to information management, combining state-of-the-art natural language processing tools to find and curate publications for review with subject matter expert, human control. This idea of a \emph{living} survey has been successfully implemented in other domains of machine learning research such as deep neural networks \cite{wang2019visual} and natural language processing research such as explainable artificial intelligence. \cite{qian2021xnlp}
While we are confident that our extensive review of the existing literature has yielded a comprehensive collection of papers,
the fast publication pace in the field and the linguistic diversity of legal research make it probable that we may have missed relevant, individual contributions.
For this survey to be as helpful for the community as possible, we have therefore built web infrastructure to accept additional contributions from the public and continuously update our results accordingly.\footnote{Available at URL: Hidden for the purpose of anonymization}
%TODO: Add URL to footnote
Researchers can provide a link to or upload their contribution, fill in a small number of fields with meta-data so that the task of maintaining the collection can be distributed across the community building on successful open source operating models.
This digital infrastructure provides us with an opportunity not only to create a living but also an \emph{interactive} survey. Since publications in the field of computational legal studies and natural legal language processing should be more than mere papers and should ideally contain both code and data for reproduction,\cite{coupette2022sharing} we have provided an easy-to-navigate graphical interface to explore the collections of papers contained in this review. Users can search for publications and filter the collection according to their needs and based on the taxonomy described in Table \ref{tab}. In addition, they can select individual and sets of publications to display temporal dynamics in methods, topics, and languages. 

\section{Conclusion and Future Perspectives}
As highlighted above, the field of Legal NLP is growing rapidly in volume and significantly in diversity of languages and sophistication of methods. As a result, research is conducted on an increasing variety of tasks as scientists push the limits of technical feasibility and demonstrate successful resolutions of increasingly difficult real-world challenges.\cite{ariai2025survey,dehghani2025large,hu2026evaluation} During our period of observation, an ever-growing amount of legal data and computational resources have become publicly available, certainly contributing to major advances in the field. However, the effects of training data, model architectures, and modeling techniques compared to the continuous increase in scale of general models require extensive further research.
Large language models have recently fully entered the public's perception, resulting in viable commercial interest from players historically disinterested in Legal NLP research, such as publishing houses, law firms, and courts. 
They often possess vast collections of data and might be increasingly willing to share them with researchers, which is evidenced by an increase in research replicability as well as increasingly sophisticated research methods combining several technological categories.
This, in turn, is likely to lead to an uptick in research using real-world, commercial, or administrative datasets.
While commercial players will drive the interest in Legal NLP research, their very own challenge will lie in the connection of the resulting, ever increasing technical capabilities into products.
Within the larger world of legal technology, language-centric technologies are therefore likely to play an ever-increasing role.
Public players will likely focus on deployment in the context of digital justice, leveraging models of the neural era to reduce case backlogs, improve access, and develop the ability to deliver justice at scale.
Our analysis has revealed that legal text generation and summarization have played a rather limited role over the first decade, but picked up speed in particular in the last few years.
As we demonstrated the usefulness of NLP methods for literature surveys themselves, one potential avenue for further research is the role of generative models for science itself and the division of labor between human researchers and large language models in our very own domain.
Future work consists of two tasks: updating the \emph{living }survey with state-of-the-art publications and building and releasing new \emph{interactive} features for exploration.
While we hope that the community can assist in the former, we plan to focus on the latter,
starting with a functionality to extract and visualize the reference graph of our collection and corresponding network metrics.
All of the above developments finally point to a growing awareness and better understanding of computational legal studies among traditional legal scholars as well as other empirically-minded scholars. This might well result in an increased intra-disciplinary collaboration and a greater openness toward quantitative methods, thereby fostering the relevance of Legal NLP in the academic community.

\section{Data Availability}
The Zenodo and Github links to data and code are hidden for the purpose of anonymization. Data and Code have been submitted with the paper. Unlike for the final publication, this includes the full text of all publications. To comply with copyright requirements, please treat the data confidential and use them only for peer review purposes.
% All papers analyzed in this study are available from conference proceedings, journals and other publication outlets. In order to respect copyright restrictions, we are not making the full texts of the papers available. However, a complete list of all papers analyzed including all metadata (e.g. authors, titles etc.) and all other preprocessed data used in this study is archived under the following DOI:

\section{Acknowledgements}
Acknowledgments are hidden for the purpose of anonymization.
%An earlier version of this work was partially funded by the German Federal Ministry of Education and Research (BMBF) through the \href{https://www.hilano.de}{HILANO} project (funding code 01IS18085).

\vspace{.1pt}

\newpage

\bibliography{bibliography}

\end{document}